\documentclass[conference]{IEEEtran}
\IEEEoverridecommandlockouts
\usepackage{cite}
\usepackage{amsmath,amssymb,amsfonts}
\usepackage{algorithmic}
\usepackage{graphicx}
\usepackage{textcomp}
\usepackage{xcolor}
\usepackage{url}
\usepackage{graphicx}
\usepackage{booktabs}
\usepackage{algorithm}
\usepackage{algorithmic}
\usepackage[switch]{lineno}
\usepackage{paralist}
\usepackage{multirow}
\usepackage{caption}
\usepackage{subcaption}
\usepackage{times}
\usepackage{soul}
\usepackage[hidelinks]{hyperref}
\usepackage[utf8]{inputenc}
\usepackage{tcolorbox}

\usepackage[T1]{fontenc}

\definecolor{lightblue}{rgb}{0.678, 0.847, 0.902}
\definecolor{lightgreen}{rgb}{0.56, 0.93, 0.56}

\def\BibTeX{{\rm B\kern-.05em{\sc i\kern-.025em b}\kern-.08em
    T\kern-.1667em\lower.7ex\hbox{E}\kern-.125emX}}
\begin{document}

\title{Sustainable Machine Learning Retraining: Optimizing Energy Efficiency Without Compromising Accuracy}

\author{
  \IEEEauthorblockN{Lorena Poenaru-Olaru\IEEEauthorrefmark{1}, June Sallou\IEEEauthorrefmark{2}, Luis Cruz\IEEEauthorrefmark{1}, Jan S. Rellermeyer\IEEEauthorrefmark{3}, Arie van Deursen\IEEEauthorrefmark{1}}
  \IEEEauthorblockA{\IEEEauthorrefmark{1}Software Engineering Research Group, \textit{Delft University of Technology}, Delft, The Netherlands \\
  \{L.Poenaru-Olaru, l.cruz, Arie.vanDeursen\}@tudelft.nl}
  \IEEEauthorblockA{\IEEEauthorrefmark{2}\textit{Information Technology Group}, \textit{Wageningen University \& Research}, Wageningen, The Netherlands \\
  \{june.sallou\}@wur.nl}
  \IEEEauthorblockA{\IEEEauthorrefmark{3}Dependable and Scalable Software Systems, \textit{Leibniz University Hannover}, Hannover, Germany \\
\{rellermeyer\}@vss.uni-hannover.de}
}


\maketitle

\begin{abstract}


The reliability of machine learning (ML) software systems is heavily influenced by changes in data over time. For that reason, ML systems require regular maintenance, typically based on model retraining.  However, retraining requires significant computational demand, which makes it energy-intensive and raises concerns about its environmental impact. To understand which retraining techniques should be considered when designing sustainable ML applications, in this work, we study the energy consumption of common retraining techniques. Since the accuracy of ML systems is also essential, we compare retraining techniques in terms of both energy efficiency and accuracy. We showcase that retraining with only the most recent data, compared to all available data, reduces energy consumption by up to 25\%, being a sustainable alternative to the status quo. Furthermore, our findings show that retraining a model only when there is evidence that updates are necessary, rather than on a fixed schedule, can reduce energy consumption by up to 40\%, provided a reliable data change detector is in place. Our findings pave the way for better recommendations for ML practitioners, guiding them toward more energy-efficient retraining techniques when designing sustainable ML software systems.




\end{abstract}



\section{Introduction}

The increasing adoption of Machine Learning (ML) and Artificial Intelligence (AI) within organizations has resulted in the development of more ML/AI software systems~\cite{greentactics}. Although ML/AI brings plenty of business value, it is known that the accuracy of ML applications decreases over time~\cite{modelmonitoringgoogle}. Thus, ML developers must monitor and maintain their ML systems in production. One reason for this phenomenon is the fact that ML applications are highly dependent on the data on which they have been trained. Real-world data usually changes over time~\cite{BAYRAM2022108632} -- a phenomenon often referred to as concept drift~\cite{conceptdriftadaptation} -- which can significantly impact the normal operation of ML systems~\cite{monitoringDataDistributionGoogle}. Therefore, appropriate maintenance techniques are required for the design of ML software systems. One common approach to maintaining these systems is to periodically update these applications by retraining the underlying ML models with the latest version of the data~\cite{datasplittingdecisions},~\cite{towardsaconsistentinterpretation}.



On another note, the process of training machine learning models has raised substantial concerns about the carbon footprint of ML applications~\cite{energyconsumptionmllifecyclegoogle, energyconsumptionmllifecyclefacebook}. For this reason, regularly retraining ML applications to preserve their accuracy, implies considerable energy consumption~\cite{mypapergreens}. On the other hand, not retraining ML applications at all can severely impact their performance over time~\cite{BAYRAM2022108632}, affecting their reliability in practice. Therefore, there is a need for sustainable ML retraining methods that reduce the environmental impact of ML applications~\cite{greentactics, mypapergreens}. 

Previous research~\cite{trinci2024greencontinuallearningreally} has explored the energy consumed during training and inference of continuous learning, which are ML applications where a machine learning (ML) model learns incrementally without requiring full retraining. Continual learning ML applications present significant challenges when incorporating domain experts' feedback into the ML model~\cite{conceptdriftadaptation}. Consequently, a continual learning approach is not suitable for certain applications. A better solution is periodically retraining ML models after accumulating sufficient new samples. However, to the best of our knowledge, the effects of retraining an ML model using various retraining techniques on both accuracy and energy consumption have not been thoroughly examined. Therefore, this study aims to examine the effects of various retraining techniques on accuracy and energy consumption in real-world ML applications, to provide best practices for designing sustainable ML systems. We analyze these retraining techniques through the lens of two perspectives, namely the \textit{data perspective} and the \textit{frequency perspective}~\cite{mypapergreens}. The \textit{data perspective} addresses the impact of the data included in the retraining process, while the \textit{frequency perspective} focuses on the impact of the frequency at which a model is retrained. Moreover, since some ML applications also require substantial energy for inference~\cite{costAIDeployment}, we also explore whether the retraining strategy affects the energy consumption of the inference tasks.


In this empirical study, we use failure prediction applications as the case study ML application. Failure prediction applications are part of the AIOps research domain and are ML applications that aim to identify failures in large and complex software systems in order to improve their efficiency and reliability~\cite{aiopschallenges}. We specifically select failure prediction applications since periodic retraining is commonly applied to keep models up to date and these applications require large datasets, making them considerably energy intensive in practical settings~\cite{datasplittingdecisions, towardsaconsistentinterpretation, modelmaturity, nodefailurepredictionconceptdrift, predictnodefailureincloude}. We argue that an efficient model monitoring and retraining strategy can lead to significant improvements in maintaining these models. Hence, we employ three failure prediction models previously presented in the literature, which are built using three open-source real-world datasets.

Previous research has compared retraining failure prediction models using all available data versus only the most recent data, with both approaches maintaining performance over time~\cite{towardsaconsistentinterpretation}. However, retraining on all data is likely to consume more energy~\cite{mypapergreens, juneluispaper}. Our study empirically evaluates the energy consumption and compares the accuracy and energy efficiency of these two retraining techniques.

Related work also suggests that periodic retraining is less sustainable compared to drift-based approaches~\cite{mypapergreens, greentactics}. However, drift detectors consume energy~\cite{omarpaper} and can be overly sensitive~\cite{mypaper}, triggering unnecessary retraining and increasing energy consumption. Hence, in this work, we quantify the impact of drift detection-based retraining on the energy efficiency of AI-driven failure prediction systems.

The main contribution of this study is a quantitative empirical study of the impact of different retraining techniques on the trade-off between accuracy and energy consumed by ML systems during both training and inference. Our research shows that drift-based retraining approaches reduce the energy consumption of failure prediction models over time. However, the gains in energy efficiency are highly dependent on the choice of a drift detector. All experiments and analyses are openly available in a replication package\footnote{\url{https://github.com/LorenaPoenaru/green_AI_maintenance}}.

\section{Background}
In this section, we introduce the background knowledge on which the remainder of the paper builds, i.e., AIOps and failure prediction, concept drift, and retraining approaches.
\subsection{AIOps}
Large-scale software systems generate vast amounts of operational data, making manual analysis and inspection impractical~\cite{aiopschallenges}. The term AIOps, introduced by Gartner~\cite{gratner}, refers to applying ML techniques to automate this process~\cite{datasplittingdecisions}. AIOps applications are ML applications used to monitor large systems and enhance software delivery, compliance, quality, and security in organizations~\cite{aiopschallenges}. A recent survey identifies four key AIOps categories: root-cause analysis, incident detection, failure prediction, and automated actions~\cite{salesforceai}. This work focuses on failure prediction, specifically analyzing two ML applications: disk failure prediction and job failure prediction, which learn patterns from past failures to anticipate hardware failure (disk) or software failure (job). It has been previously shown~\cite{datasplittingdecisions, diskfailureprevwork, diskfailureprevwork1, towardsaconsistentinterpretation, modelmaturity} that these applications are severely affected by concept drift and that they employ large amounts of data to train. 

\subsection{Concept Drift}
The term concept drift refers to changes in the data over time~\cite{learningUnderConceptDrift}. It is a ubiquitous phenomenon in real-world data, as data changes are generated by uncontrollable external factors~\cite{mypaper}. Concept drift can severely impact the accuracy/performance of ML models over time since the ML algorithms used to build these models work under the assumption that the distribution of the data data learned during the training process should be similar to the distribution of the data on which the model is evaluated~\cite{BAYRAM2022108632}. However, in the real world, this assumption often does not hold, since data are continuously changing, which can lead to noticeable drops in the model's accuracy over time.


Continuous model update/retrain is a commonly known technique to mitigate the effects of concept drift on ML models over time~\cite{conceptdriftadaptation, learningwithdriftdetection, mitigatingconceptdrift}. 

From the \textbf{retraining frequency perspective} researchers~\cite{conceptdriftadaptation} have distinguished between two retraining techniques, namely \textit{periodic retraining} and \textit{informed retraining}. Periodic retraining implies that models are retrained based on a predefined period, while informed retraining means that a model is retrained based on a data monitoring tool called a concept drift detector. In the latter situation, the concept drift detector evaluates whether the training data becomes significantly different than the data the model is evaluated on~\cite{mypaper}. 

From the \textbf{retraining data perspective}, there are two techniques previously presented in the literature, namely the \textit{sliding window} and the \textit{full-history} retraining approach~\cite{towardsaconsistentinterpretation, modelmaturity, mypaperanomalydetection}. The sliding window approach implies that the model is retrained only on the newest data, discarding old samples, while with a full-history retraining approach, the model is retrained on all the available data until a certain point in time. Therefore, the latter retraining technique constantly enriches the training dataset with new samples once they become available.

\section{Related Work}
\subsection{Concept Drift in AIOps Applications}

Previous work has shown that multiple ML applications within AIOps, including failure prediction model, have been affected by concept drift~\cite{aiopschallenges, datasplittingdecisions, nodefailurepredictionconceptdrift, predictnodefailureincloude, mypaperanomalydetection, issreconceptdrift}. In failure prediction applications, concept drift can be caused by different external factors, such as feature updates, user workloads, or software/hardware updates~\cite{nodefailurepredictionconceptdrift}. These factors can affect the behavior of the data over time, which usually impacts the accuracy of failure prediction models. This aspect makes concept drift a serious threat to the trustworthiness of failure prediction models, especially when their output is used in decision-making processes~\cite{aiopschallenges}.

When it comes to AIOps applications, previous works recommend AIOps practitioners to periodically retrain failure prediction models~\cite{nodefailurepredictionconceptdrift, datasplittingdecisions, towardsaconsistentinterpretation}. From the retraining data perspective, both the sliding window~\cite{datasplittingdecisions, towardsaconsistentinterpretation, modelmaturity, slidingwindowretrain, predictnodefailureincloude, nodefailurepredictionconceptdrift, mypaperanomalydetection} and full-history approaches~\cite{towardsaconsistentinterpretation, modelmaturity, mypaperanomalydetection} retraining approaches were employed to update AIOps models over time. From the retraining frequency perspective, the most commonly used retraining technique is the periodic retraining~\cite{datasplittingdecisions, towardsaconsistentinterpretation, modelmaturity, nodefailurepredictionconceptdrift, predictnodefailureincloude}. However, Lyu et al.~\cite{modelmaturity} also experimented with informed retraining. The authors employed a supervised concept drift detector to monitor the error of the failure prediction model over time and retrained every time they observed a significant error increase. To compute the error, the ground truth, also known as true labels, is required. True labels refer to which sample is a "failure" and which sample is a "non-failure". However, in some AIOps applications, continuously monitoring the error over time might not be possible since obtaining true labels is expensive in time and resources. For these applications, to obtain the true labels operational engineers have to continuously perform root cause analysis to understand the main cause of the failure\cite{aiopschallenges, mypaperanomalydetection, issreconceptdrift}. This consumes a significant amount of the time of operational engineers which makes acquiring the true labels significantly expensive. In this situation, employing an unsupervised drift detector that does not require computing the error using true labels would be preferred~\cite{mypaperanomalydetection, mypaper}. Therefore, unlike previous work~\cite{modelmaturity}, in this paper, we analyze the effects of retraining based on an unsupervised drift detector. Furthermore, while previous work only took into account the accuracy when determining the effects of retraining based on drift detection, we also consider the consumed energy to ensure the sustainability of the failure prediction models over time.

\subsection{Sustainability of ML Systems}

The adoption of ML applications in the industry has seen considerable growth over the past years~\cite{energyconsumptionmllifecyclefacebook, energyconsumptionmllifecyclegoogle}. Although numerous domains benefit from the use of machine learning, these models consume significant amounts of energy, raising concerns among researchers about the environmental impact of such applications~\cite{energyconsumptionmllifecyclefacebook, energyconsumptionmllifecyclegoogle, greentactics, costAIDeployment}. This led to the rise of the GreenAI research field, which encourages the development of ML applications that consume less energy while preserving their accuracy~\cite{greenAI}.


One key practice in building GreenAI systems is reporting the energy consumption of ML applications, which could raise awareness of their carbon footprint and help find more sustainable configurations~\cite{greenAI}. Wu et al.~\cite{energyconsumptionmllifecyclefacebook} examined the carbon footprint of Facebook’s ML applications and discovered that both training and inference have a significant contribution to the overall carbon footprint of the ML application. Plenty of research has been focused on understanding the environmental impact of training ML systems, such as the study of Xu et al.~\cite{greenAISilveiro} targeting multiple computer vision applications. On the other hand, the work of Luccioni et al.~\cite{costAIDeployment} analyzed inference energy in tasks like image classification and language modeling. They concluded that although tasks involving images are more energy-intensive, model training remains significantly more carbon-intensive compared to inference.

While significant attention has been given to the energy consumption of training and inference in ML models, the energy used during the model's active production phase (model lifecycle) has received less focus. Trinci et al.~\cite{trinci2024greencontinuallearningreally} investigated the training and inference energy efficiency of continual learning algorithms for computer vision applications, but these algorithms are not directly applicable in the AIOps context due to the high costs of continuously gathering true labels\cite{aiopschallenges} to perform continual learning. AIOps applications require ongoing model monitoring and updates, but these should occur in batches at predefined intervals rather than continuously. Although previous work~\cite{mypapergreens} explored sustainable retraining techniques, there is a lack of empirical analysis demonstrating their effectiveness in maintaining model accuracy and reducing energy consumption over time. Omar et al.~\cite{omarpaper} have studied the energy consumption of multiple drift detectors with respect to their drift detection accuracy. However, the drift detectors analyzed operate in a supervised manner, relying on error rate computation after inference, which requires true labels. This approach is impractical for some AIOps applications, where acquiring true labels depends on human annotation\cite{aiopschallenges}. Therefore, unsupervised drift detectors are more suitable, as they identify drift by analyzing differences between training and inference data without needing true labels. Our study differs from previous work~\cite{omarpaper} by employing unsupervised drift detectors and focusing on their ability to indicate the need to retrain a failure prediction model, rather than assessing the drift detection accuracy.

\section{Research Questions}
The goal of our study is to assess the impact of different retraining techniques on the energy consumption of failure prediction models. With this aim, we address the following research questions (RQ): 
\begin{compactenum}



    \item [RQ1.] What is the impact of each retraining technique on the training energy consumption?
    \begin{compactenum}
        \item What is the impact on energy consumption of employing the sliding window vs the full-history approach (retraining data perspective)?

        \item What is the impact on energy consumption of employing the periodic vs the informed retraining (retraining frequency perspective)?

        \item What is the best retraining technique overall?
    \end{compactenum}
    
    \item [RQ2.] What is the impact of each retraining technique on the inference energy consumption? 

\end{compactenum}

\section{Methodology and Experiments}
In this section, we outline the methodology used to address the research questions and conduct our experiments. Specifically, we discuss the used datasets, the process for building the failure prediction models, the experimental design, and the energy consumption measurement.

\subsection{Datasets}
We employ the only three publicly available open-source AIOps datasets to build failure prediction models according to previous work~\cite{datasplittingdecisions, modelmaturity}: the Backblaze Disk Stats Dataset~\cite{backblazedata}, the Google Cluster Traces Dataset~\cite{googleclusterdata}, and the Alibaba GPU Cluster Trace Dataset~\cite{alibabadata}. 

The \textbf{Backblaze Disk Stats Dataset} has been previously used to build disk failure prediction models~\cite{datasplittingdecisions, diskfailureprevwork1, towardsaconsistentinterpretation, diskfailureprevwork, modelmaturity}. It contains information about operational hard drives available in the data center collected daily since 2013. The dataset includes both drive information (manufacturer, serial number, or capacity), as well as information related to early error detection extracted through a monitoring system implemented by the manufacturer, called SMART attributes (Self-Monitoring, Analysis, and Reporting Technology). Given that data collected before 2015 does not include SMART attributes and to align with previous work~\cite{datasplittingdecisions}, we employ 12 months of data collected in 2015.

The \textbf{Alibaba GPU Cluster Trace Dataset} is relatively new, released in 2021 by the Alibaba Group, and, to the best of our knowledge, there has been only one work that employed it to build job failure prediction models~\cite{modelmaturity}. It is composed of information regarding job execution extracted from a large-scale data center. The data is collected for a period of 2 months, from July to August 2020~\cite{mlaas}, from approximately 6500 GPUs across around 1800 machines. 

The \textbf{Google Cluster Traces Dataset} has been previously used to build job failure prediction models~\cite{googletraceprediction1, datasplittingdecisions, towardsaconsistentinterpretation, modelmaturity}. The dataset includes information about jobs executed on a large-scale cluster at Google collected for 29 days in May 2011.

\subsection{Machine Learning Models}

When building the failure prediction models, i.e., disk failure prediction for the Backblaze dataset and job failure prediction for the Alibaba and Google datasets, we replicate the pipelines presented in previous work~\cite{datasplittingdecisions, towardsaconsistentinterpretation, modelmaturity, diskfailureprevwork, diskfailureprevwork1, googletraceprediction1}. Thus, we use the same features and follow the same model design strategy presented in previous studies. More details about features are specified in our replication package.

\subsubsection{Labels and Features}

The Backblaze dataset contains an attribute indicating whether a drive failed the day after data collection, which serves as the label for our failure prediction model. It consists of approximately 7 million samples, with 0.05\% labeled as failures and 99.95\% as non-failures. For the prediction model, we selected 19 temporal SMART features from previous research \cite{datasplittingdecisions, modelmaturity, diskfailureprevwork1, diskfailureprevwork, towardsaconsistentinterpretation}. Of these, 11 are non-cumulative (raw) values from the last day, while 8 are cumulative (raw diff) changes over one week compared to the previous day's values.

To create the final Alibaba dataset, we remove unfinished jobs and those ending in less than five minutes, except those labeled with the status "fail." The final dataset consists of 701,000 samples, with 34.5\% labeled as failures and 65.5\% as non-failures. To train the job failure prediction model, we use 12 features, comprising 6 configuration (conf) features and 6 temporal (temp) features calculated over a 5-minute window since job submission.

The Google dataset lacks a direct attribute for failed or non-failed jobs. Each job (sample) can have multiple events (fail, finish, kill, submit, update, evict, schedule) and states (pending, dead, running, unsubmitted). Following previous research~\cite{datasplittingdecisions, towardsaconsistentinterpretation, modelmaturity}, a job is labeled as "fail" only if its final state is "fail." We remove jobs with incomplete records and those that finished within five minutes of submission, as they do not provide sufficient metrics for predicting failure. The final dataset includes approximately 625,000 samples, with 1.5\% labeled as failures and 98.5\% as non-failures. For the job failure prediction model, we use 15 features: 9 configuration (conf) features and 6 temporal (temp) features calculated over a 5-minute period since job submission.

\subsubsection{Model Building Pipeline}
\begin{figure}
    \includegraphics[width=0.45\textwidth]{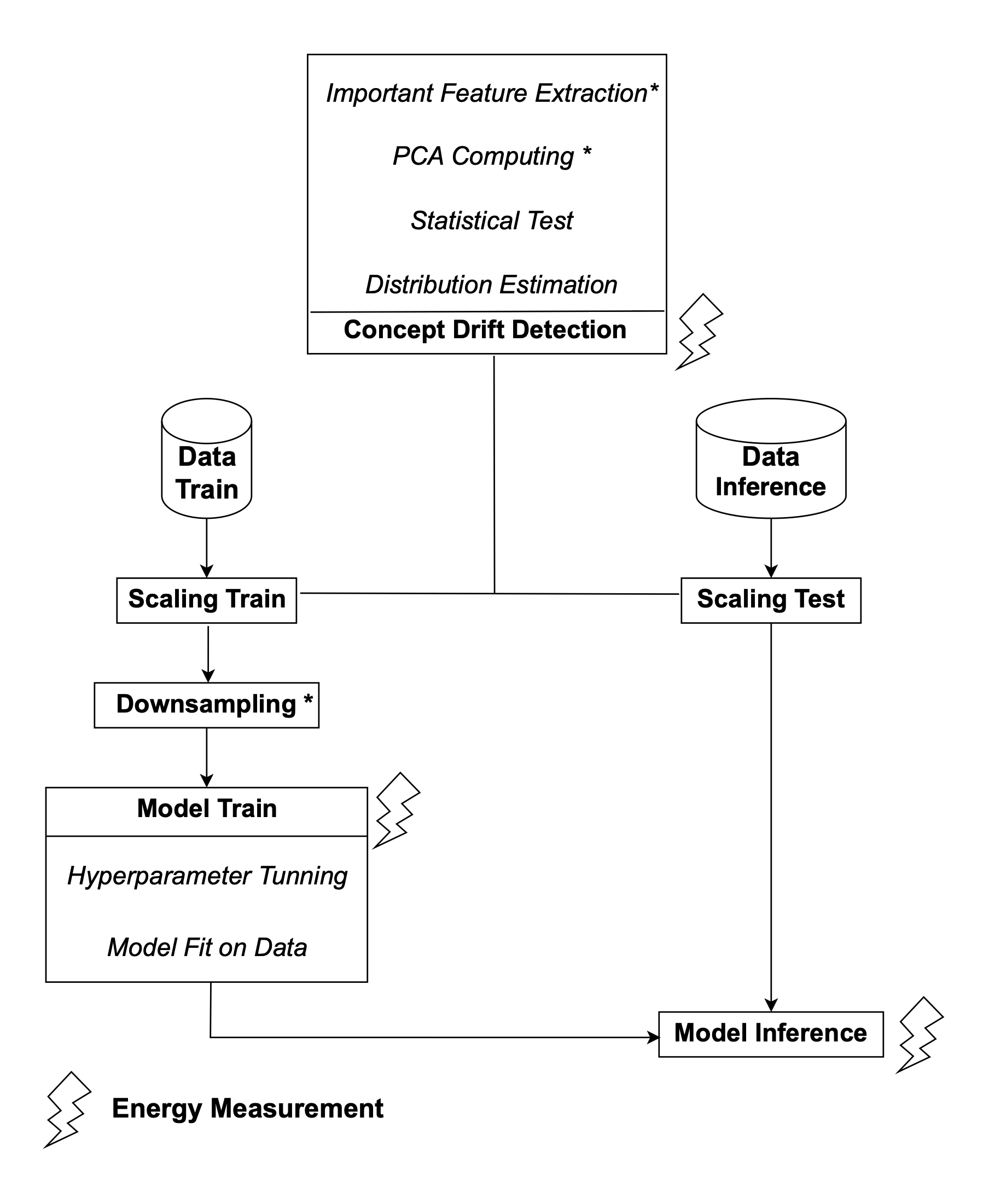}
    \caption{Energy measurements in model pipeline. The symbol `*' indicates that the specific step was done only when applicable.}
    \label{figure:model_pipeline_measurements}
    \vspace{-1em}
\end{figure}

We built three model failure prediction models, one disk failure prediction, and two job failure prediction models, corresponding to the three datasets employed in this study, Backblaze, Google, and Alibaba, respectively. To closely resemble the failure prediction models presented in previous work~\cite{datasplittingdecisions, towardsaconsistentinterpretation, diskfailureprevwork, diskfailureprevwork1, modelmaturity, googletraceprediction1} we build a monthly failure prediction model for the Backblaze dataset, a daily failure prediction model for the Google dataset and a weekly failure prediction model for the Alibaba dataset. We split each dataset in half. The first part is used to train each model, while the second part is further divided into smaller subsets, each corresponding to a specific period (day, week, or month, depending on the dataset). These subsets mimic the real-world scenario in which new data is generated periodically and are used for inference. Fig.~\ref{figure:model_pipeline_measurements} shows a detailed overview of our model-building pipeline.

The model-building pipeline is similar for all our three failure prediction models unless specified otherwise. The first step of the pipeline is preprocessing the data through scaling. Similar to previous work~\cite{datasplittingdecisions, towardsaconsistentinterpretation, modelmaturity}, we employ a standard scaler which we fit on the training data and apply on the inference data such that all values of the features range in a predefined interval. 

The Backblaze and Google datasets suffer from high-class imbalance, meaning that the number of failure samples is tremendously lower compared to the number of non-failure samples. For the Alibaba dataset, the class imbalance is not as severe. Therefore, similar to previous work~\cite{datasplittingdecisions, modelmaturity, towardsaconsistentinterpretation}, we perform downsampling for the Backblaze and Google datasets obtaining a class balance ratio of 1:10. This indicates that for each failure sample, there are 10 non-failure-samples.

The last steps of the model-building pipeline are model training and inference. For this step, we employ a Random Forest classifier as it is one of the most popular classifiers used in failure prediction achieving the highest performance in distinguishing between failure and non-failure samples~\cite{towardsaconsistentinterpretation, diskfailureprevwork, diskfailureprevwork1, datasplittingdecisions, modelmaturity}. While training the model, hyperparameter tuning is performed through Randomized Search, similar to previous work~\cite{modelmaturity}. The trained model is further used for inference.

\subsubsection{Failure Prediction Evaluation Metric}
Similar to previous work~\cite{datasplittingdecisions, towardsaconsistentinterpretation, modelmaturity}, we employ ROC AUC to evaluate the accuracy of the failure prediction model. In this paper, we will use the terms ROC AUC and accuracy interchangeably. ROC AUC is a metric that measures the model's ability to assign higher predicted probabilities to the positive class compared to the negative class. The highest ROC AUC score is 1.0 and shows that all samples are correctly classified. We choose this metric due to the high-class imbalance in all datasets.

\subsection{Retraining Approaches}
In our experiments, we investigate the energy consumption of different retraining techniques. From the retraining data perspective, we analyze the effects of two retraining techniques, namely the \textit{full-history approach} and the \textit{sliding window approach}~\cite{towardsaconsistentinterpretation}. 
In full-history retraining, the model is retrained using all data accumulated up to the given retraining time.  
In sliding window retraining, at each retraining step, new data is incorporated while the oldest data, corresponding to one period, is discarded. Unlike full-history retraining, this method maintains a relatively constant training dataset size.

From the retraining frequency perspective, we analyze the effects of two retraining techniques, namely the \textit{informed} vs \textit{periodic} retraining~\cite{conceptdriftadaptation}.
In the periodic retraining approach, the model is retrained at predefined intervals, regardless of whether the data distribution has changed. To remain consistent with prior work \cite{modelmaturity, datasplittingdecisions, towardsaconsistentinterpretation}, we perform retraining at different intervals depending on the dataset: monthly for Backblaze, weekly for Alibaba, and daily for Google.
In contrast, the informed retraining approach triggers model updates only when a significant shift in the data distribution is detected. To identify such shifts, we employ the Kolmogorov-Smirnov (KS) statistical test \cite{BAYRAM2022108632, mypaper, learningUnderConceptDrift, augur}, a widely used unsupervised drift detection technique.

To further analyze the impact of different drift detection strategies on energy consumption, we explore multiple variations of drift detection. This process consists of two primary steps: first, estimating the distributions of both the training and inference data, and second, applying the KS test to compare these distributions. We examine three distinct approaches to drift detection. The first approach, KS-ALL, estimates distributions using all available model features. The second, KS-PCA, reduces the feature dimensionality such that 95\% of the variance in the data is preserved using Principal Component Analysis (PCA) before estimating distributions. The third approach, KS-FI, selects only the most relevant features, filtering out those with a feature importance ranking below the mean, as determined by Gini importance. Both KS-PCA and KS-FI introduce a feature selection step before distribution estimation. While KS-PCA applies PCA for dimensionality reduction, KS-FI retains only the most informative features based on their importance ranking.

\subsection{Experimental Design \& Setting}
In our experimentation, we study 8 retraining configurations based on two retraining perspectives: data and frequency.

We repeat our experiments to mitigate bias in our results caused by the randomness of the ML model. We allocate a one-week budget for each dataset, running each configuration multiple times. For the Backblaze and Alibaba datasets, we repeat the experiments 30 times using 30 different random seeds. The Google dataset is one of the most computationally intensive, therefore only 5 random seeds are computed. However, no significant variation in the results was observed by experimenting with different random seeds. Furthermore, when collecting energy consumption data, we shuffle the configurations to reduce the risk of background activities or factors such as temperature impacting only a category of experiments, following recommendations for energy studies~\cite{cruz2021green}. All our conclusions are verified using the Wilcoxon statistical test with a confidence level of 95\%.

The experiments are run on a machine with an AMD Ryzen 9 7900X processor (12 physical cores, 24 threads), 64 GB RAM (2x32GB DDR5 @ 5.600MT/s), and an MSI Geforce RTX 4090 (24GB GDDR6X memory) graphic card. The operating system is Ubuntu 22.04.3, with Linux kernel version 6.2.0.

\subsection{Energy Consumption}
We measure energy at the level of the \textit{model pipeline} and the \textit{model lifecycle}. The model pipeline level refers to the components of the failure prediction model that are measured. The model lifecycle refers to how the total energy consumed by a certain process is consumed throughout the lifecycle of the machine learning model. In this subsection, we further describe which tool we employ to measure energy consumption.

\subsubsection{Energy Measurement Tool}
Energy measurement tools help quantify the sustainability of different experimental settings. To measure the consumed energy, we employ the CodeCarbon~\cite{codecarbon} Python package. The energy consumed by the CPU and RAM is measured through RAPL and the energy consumed by the GPU is measured through NVIDIA Management Library~\cite{greenAISilveiro}. CodeCarbon measures the duration of each experiment (in seconds) and the consumed energy (in kWh). Furthermore, this energy measurement tool has been previously used to measure the energy consumed by drift detectors and the energy consumed during training and inference~\cite{omarpaper, trinci2024greencontinuallearningreally}.


\subsubsection{Measuring Energy Consumption of the Model}
In our experiments, we measure the energy consumption of multiple steps in the failure prediction model lifecycle as shown in Figure~\ref{figure:model_pipeline_measurements}. In this work, we are solely interested in understanding the impact of different model retraining techniques on the energy consumed during the model lifecycle. Thus, we are not measuring the energy of all the steps required in building the models, such as data scaling or downsampling to balance the two classes. In our experiments, we measure three types of energy, namely \textit{training energy}, \textit{drift detection energy} and \textit{inference energy}. Furthermore, in our results, we depict the cumulative values of the energy of these models. For example, the training energy is composed of the initial training of the model and energy consumed to perform retraining either periodically (if the configuration is Periodic) or when the drift detection indicates (if the configuration is drift detection based). For the Static configuration, we perform training only once and never retrain further.

The \textbf{training energy} is the total energy consumed by training/retraining. This energy measurement is composed of the energy measured during the hyperparameter tuning phase, in which the best hyperparameters are chosen to fit the training data, and the energy measured during the phase in which the best model is fitted on the training data as described in Fig.~\ref{figure:model_pipeline_measurements}.

\textbf{Drift detection energy} refers to the total energy consumed in detecting data drift. Depending on the detector—KS-ALL, KS-PCA, or KS-FI—this energy is measured in two or three components. KS-ALL includes energy used for estimating the data distribution and applying a statistical test to determine significant changes. KS-PCA adds a third measurement for the energy spent on dimensionality reduction using PCA. Similarly, KS-FI requires three measurements, but instead of PCA, it measures the energy used to extract and filter important features.


The \textbf{inference energy} represents the energy consumed by applying the trained model to perform inference and extract the predictions.

\section{Results}
In this section, we present our results and answer the research questions.


\subsection{Impact on Training Energy}

To answer RQ1, we depict our results in Fig.~\ref{figure:results_violin_plots}, presenting at the top the results of ROC AUC and at the bottom the total training energy combined with the total drift detection energy for each retraining technique for each dataset, when applicable. In the case of the Static model, the total training energy contains only the initial model training. When it comes to the Periodic model, there is no total drift detection energy, thus we solely depict the energy consumed during the periodical model retraining process. The drift detection energy is solely considered for the KS-ALL, KS-FI, and KS-PCA models.

\subsubsection{RQ1a: Full-History vs Sliding Window Retraining}
In this experiment, we evaluate the sliding window and full-history approaches in terms of model accuracy (ROC AUC) and energy consumption (joules). To enhance the generalizability of our conclusions, we assess these retraining strategies in two scenarios: periodic retraining (Periodic) and drift-based retraining using the KS-ALL, KS-FI, and KS-PCA methods. Additionally, we include a baseline scenario with a static model (Static), trained only once, to assess the overall benefits of retraining.


\begin{figure*}
    \centering
    \includegraphics[width=1.0\textwidth]{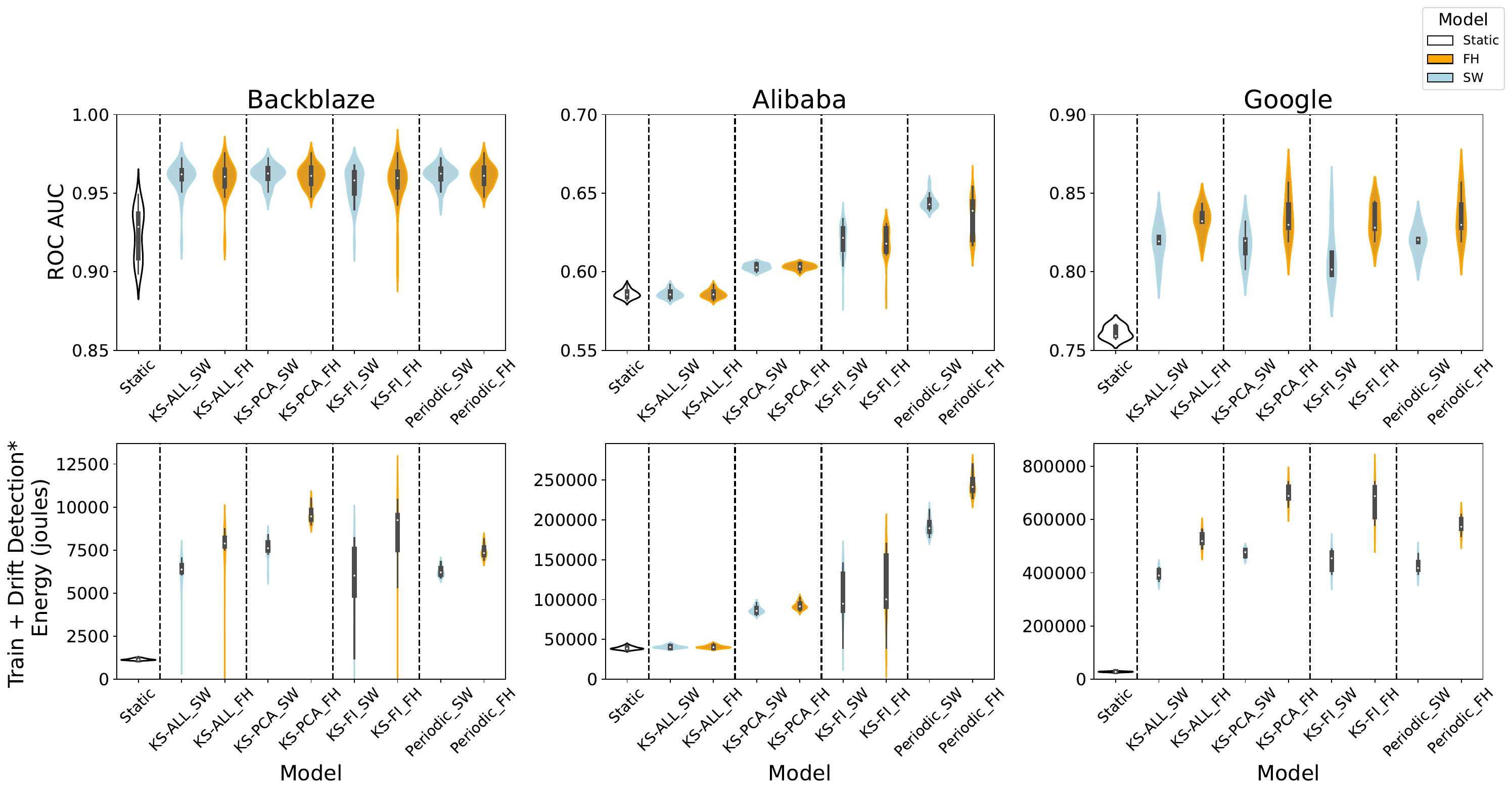}
    \caption{ROC AUC and energy consumption of each retraining technique. Note: for the informed retraining (KS-ALL, KS-PCA, and KS-FI all configurations) the energy consumed is calculated during both the retraining and the drift detection.}
    \label{figure:results_violin_plots}
    \vspace{-1em}
\end{figure*}


In Fig.~\ref{figure:results_violin_plots} we show that in almost all situations the full-history approach is more energy-intensive when compared to the sliding window approach. The only exception can be observed in the situation of KS-ALL for the Alibaba dataset, where there is no significant difference (p-value of 0.67) in terms of the amount of energy consumed using a sliding window or a full-history retraining approach. This exception can be explained by the fact that the ROC AUC for KS-ALL\_FH is similar to the ROC AUC for KS-ALL\_SW and to the ROC AUC of the Static model. This shows that in this particular situation for both full-history and sliding window approaches the drift detector did not identify any drift, and, therefore, no retraining has been performed besides the initial model training. In all other situations, the energy consumed by retraining using a sliding window approach was statistically lower than the energy consumed by retraining using a full-history approach (p-values below 0.05). For instance, when it comes to the Periodic model, we can notice an almost 25\% decrease in energy consumed while retraining using the sliding window approach vs the full-history approach.

\begin{tcolorbox}[title=RQ1a Answer, colback=lightblue!5]
A sliding window approach consumes significantly less energy than a full-history approach.
\end{tcolorbox}

\subsubsection{RQ1b: Periodic vs Informed Retraining}
From Fig.~\ref{figure:results_violin_plots} we can notice that in most cases the train energy consumption is reduced when using an informed retraining (retraining based on drift detection) technique for all datasets.

When it comes to the Alibaba dataset, the energy consumed by training based on the KS-ALL drift detector is significantly similar to the energy consumed in the Static configuration. This shows that the KS-ALL drift detector does not identify any change in the data, and thus, any need for retraining, consuming almost the same amount of energy as the Static model. However, there is a significant difference in the ROC AUC between KS-ALL and the other configurations (at least 4\%), showing that retraining is required. Thus, this drift detector is not able to properly identify the drift in this dataset, and should not be used as an indicator of when to retrain. We can also notice that some informed retraining techniques consume more energy than periodic retraining techniques. For instance, retraining a model for the Google dataset using a KS-FI or a KS-PCA drift detector is more energy-intensive than periodically retraining the model. The reason for this is the fact that these drift detectors are too sensitive to changes in data and signal the need to retrain often. Since these detectors also consume energy and the number of times the model required retraining is not significantly reduced, these configurations consume more energy compared to periodic retraining. On the other hand, for the Google dataset, we can observe that some drift detectors, KS-FI and KS-PCA, are considerably sensitive and constantly indicate the need for retraining. For this reason, they consume more energy compared to periodic retraining and this finding shows that informed retraining is not always more sustainable than periodic retraining.

\begin{table}[ht]
\centering
\caption{Percentage of energy consumed by each drift detector from the energy consumed for training + drift detection.}
\begin{tabular}{| c | c | c | c | c | c | c | c | c | c |} 
 \cline{1-5}
  \multicolumn{1}{|c|}{\textbf{Drift Detector}} &
 \multicolumn{1}{|c|}{\textbf{Retraining}}& 
 \multicolumn{1}{|c|}{\textbf{Backblaze}}& 
 \multicolumn{1}{|c|}{\textbf{Alibaba}}& 
 \multicolumn{1}{|c|}{\textbf{Google}}\\
 
\cline{1-5}
KS-ALL & SW  & 3.17 & 3.90 & 0.41 \\
\cline{2-5}
& FH  & 2.65 & 3.91 & 0.37  \\
 \cline{1-5}
KS-PCA & SW  & 1.77 & 1.54 & 0.26 \\
\cline{2-5}
 & FH  & 1.50 & 1.53 & 0.21  \\
\cline{1-5}
 KS-FI & SW  & 1.63 & 0.47 & 0.13  \\
 \cline{2-5}
 & FH  & 1.16 & 0.47 & 0.11  \\
 \cline{1-5}
\end{tabular}
\label{table:percentage_drift_detection}
\end{table}

In Table~\ref{table:percentage_drift_detection}, we present the percentage of the energy consumed solely by the drift detector out of the energy consumed by both training and drift detection. This table shows that the energy consumed by the drift detector is relatively low (less than 4\%). This shows that incorporating an unsupervised drift detector does not bring too much overhead to the total energy consumption. We can further see from Table~\ref{table:percentage_drift_detection} that the KS-ALL drift detector consumes more energy than KS-PCA and KS-FI for all datasets. This can be explained by the fact that this drift detection technique derives the data distribution from the entire feature space, while the KS-PCA and KS-FI perform feature reduction before deriving the data distribution. Thus, unsupervised drift detectors incorporating feature reduction are generally sustainable. 

\begin{tcolorbox}[title=RQ1b Answer, colback=lightblue!5]
Informed retraining usually consumes less energy than periodic retraining when the drift detector is properly chosen.
\end{tcolorbox}

\subsubsection{Best Overall Retraining Approach}
When choosing the best overall retraining approach, we have to consider which retraining technique achieves the highest ROC AUC while lowering energy consumption compared to periodic retraining. Furthermore, we consider the Static model as a lower bound in terms of ROC AUC.

\begin{figure*}
    \centering
    \includegraphics[width=1.0\textwidth]{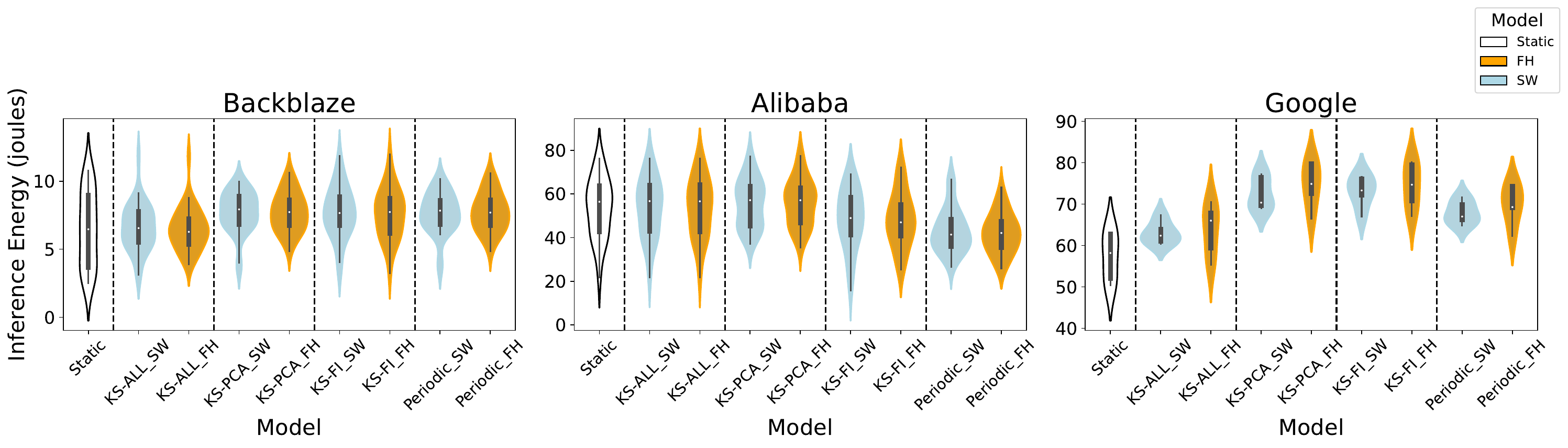}
    \caption{Energy consumed during inference for each retraining technique.}
    \label{figure:results_violin_plots_inference}
    \vspace{-1em}
\end{figure*}

To answer this research question (RQ1c), we examine the energy consumed for each retraining technique with respect to its corresponding ROC AUC improvement depicted in Figure~\ref{figure:results_violin_plots}. When it comes to the Backblaze datasets, there is no statistical difference in ROC AUC among configurations that include a drift detector (KS-ALL, KS-PCA, and KS-FI) and the periodic configuration. However, the KS-FI using a sliding window approach consumes the least energy. The KS-FI\_SW is the best overall for the Alibaba dataset, achieving the highest ROC AUC with the lowest energy use. Regarding the Google dataset, the retraining technique that consumes the least amount of energy while achieving the highest performance is the KS-ALL drift detector. These experiments demonstrate that no single retraining technique is universally optimal. Therefore, further experimentation is necessary to identify the most sustainable approach for each case.

\begin{table}[ht]
\centering
\caption{Energy estimation for using a drift detection-based retraining approach vs a periodic approach over the period of one . In the Drift Detection column, we depict the most sustainable drift detection-based retraining technique for each dataset, namely KS-FI for Backblaze and Alibaba and KS-ALL for Google. Note: Only the sliding window retraining technique is presented since it is the most sustainable. In this table, K implies that the value needs to be multiplied by 1.000 and M implies that the value needs to be multiplied by 1.000.000}
\begin{tabular}{| c | c | c | c | c | c | c | c | c |} 
 \cline{1-4}
  \multicolumn{1}{|c|}{\textbf{Drift Detector}} &
 \multicolumn{1}{|c|}{\textbf{Retraining}}& 
 \multicolumn{1}{|c|}{\textbf{Periodic}}& 
 \multicolumn{1}{|c|}{\textbf{Drift Detection}}\\
 
\cline{1-4}
Backblaze & 6 months  & 6.3K &  5.9K \\
\cline{2-4}
& 1 year estimate  & 12.6K & 11.8K \\
 \cline{1-4}
Alibaba & 1 month  & 191.3K & 106.9K \\
\cline{2-4}
 & 1 year estimate  & 2.3M & 1.3M  \\
\cline{1-4}
 Google & 2 weeks & 427.0K & 393.7K \\
 \cline{2-4}
 & 1 year estimate  & 11.1M & 10.2M \\
 \cline{1-4}
\end{tabular}
\label{table:one_year_estimation}
\end{table}

Each of our datasets was evaluated during periods of different lengths, namely 6 months for Backblaze, 1 month for Alibaba, and 2 weeks for Google. These lengths are strongly dependent on the size of each dataset and on how failure prediction models were designed in previous work~\cite{datasplittingdecisions, towardsaconsistentinterpretation, modelmaturity}. Therefore the retraining frequency for each failure prediction model is different, monthly retraining for Backblaze, weekly retraining for Alibaba and daily retraining for Google. To better understand the impact of retraining periodically vs retraining based on drift detection over a longer period, we estimated the energy consumed on training + drift detection over one year for each dataset. In this experiment, we assume no change in the energy consumed during the given period and the rest of the year. For instance, if the energy consumed by periodic retraining during the available 6 months for Backblaze is 6.3K, we assume that in the following 6 months, the model will consume the same amount of energy.

In Table~\ref{table:one_year_estimation} we show the energy consumed by the most sustainable retraining technique for each dataset (KS-FI for Backblaze and Alibaba and KS-ALL for Google) vs periodic retraining and an estimate of energy consumed over one year. Our results show that retraining using a suitable drift detector can significantly reduce energy consumption over one year. More specifically it can reduce the energy consumed by failure prediction models for Google and Alibaba by around 1000 kilojoules (around 40\% less energy for Alibaba and 10\% for Google) and the energy consumed by failure prediction models for Backblaze by almost 1 kilojoule (around 7\%). Therefore, employing a drift detector that is not too sensitive and can identify when a model requires retraining is substantially beneficial in the long term compared to periodic retraining.

\begin{tcolorbox}[title=RQ1c Answer, colback=lightblue!5]

There is no one-size-fits-all solution for model retraining and different options should be considered before opting for a particular technique. Although there is evidence of the benefits of using an informed retraining technique, the main challenge lies in choosing an appropriate drift detector and in the frequency of concept drift occurrences. Furthermore, the optimal informed retraining technique should be paired with a sliding window retraining approach to minimize the consumed energy.
\end{tcolorbox}

\subsection{Impact on Inference Energy}
Our study aims to understand whether different model retraining techniques can impact the energy consumed during inference. Therefore, in Fig.~\ref{figure:results_violin_plots_inference} we depict the distribution of the energy consumed during inference for each model across all random seeds. From Fig.~\ref{figure:results_violin_plots_inference} we can notice that the variation in inference energy consumption for each retraining technique is relatively small. Therefore, although each retraining technique is different in terms of the frequency of retraining a model and the data, the energy on inference is not impacted. 

We verified whether there is a significant difference between the energy consumed during the inference for all analyzed scenarios (p-values higher than 0.05). For the Google dataset, there was no significant difference in energy consumed during inference between all the analyzed retraining techniques. For the Alibaba dataset, both periodic retraining techniques (Periodic\_SW and Periodic\_FH) are significantly different from the rest in terms of energy consumed during inference. However, the difference in consumed energy is only around 10 joules. When it comes to Backblaze, there was no significant difference in the energy consumed during inference between the Static and KS-ALL or between KS-PCA, KS-FI, and Periodic models. However, although for instance the Static and Periodic models consume significantly different amounts of energy according to the Wilcoxon statistical test (p-value 0.04), this difference was only around 1.5 joules. Thereby, the difference in energy consumed during inference between the retraining techniques is either not significant or extremely low.

\begin{tcolorbox}[title=RQ2 Answer, colback=lightblue!5]
In general, the employed retraining technique does not affect the energy consumption of inference tasks.
\end{tcolorbox}

\section{Discussion and Implications}

In this section, we will highlight the general findings derived from this study and we will discuss our results. Our findings aim to help ML practitioners build more sustainable ML software systems and understand how to experimentally assess different retraining techniques in terms of sustainability. The remainder of this section will present each of our findings followed by a discussion and its implications.
 
\begin{tcolorbox}[title=Finding 1, colback=lightgreen!5]
Retraining using only the newest data is typically more sustainable than retraining on all available data and has a negligible influence on the model's performance.
\end{tcolorbox}

In our work, we validate the claim of the authors~\cite{mypapergreens} that employing a sliding window approach when retraining ML models is more sustainable with experimental evidence. When comparing the sliding window and the full-history approach we noticed that for all datasets the difference in model performance, ROC AUC, is minimal, while the difference in energy consumption is considerable. For the Backblaze dataset, the ROC AUC obtained by both retraining approaches is the same, while for the Alibaba dataset, the sliding window approach achieved an ROC AUC of 1\% higher than the full-history approach, showing that the model accuracy benefits from deleting older samples. The only dataset where the full-history approach has a higher ROC AUC than the sliding window approach is the Google dataset. However, from all three datasets, the Google dataset is the shortest in terms of sample collection (29 days instead of 2 months for Alibaba and one year for Backblaze). Therefore, the reason why a full-history approach can be more beneficial from the accuracy perspective is that the model might require more training samples before deploying the model in production. Furthermore, the difference in energy consumed by retraining the Google dataset with a full-history approach vs a sliding window approach is significantly high, ranging from 130 kilojoules to 222 kilojoules depending on the configuration, while the ROC AUC gain is only a maximum of 1.5\%. Due to the minimal accuracy improvement and the high energy consumption associated with using a full-history approach, we recommend that ML practitioners adopt a sliding window retraining technique when developing Green ML applications.

\begin{tcolorbox}[title=Finding 2, colback=lightgreen!5]
Retraining a model based on a drift detector can benefit both the model's performance and the energy consumed \textbf{only} if the drift detector is properly chosen.
\end{tcolorbox}

While answering RQ1b and RQ1c, we noticed that integrating a drift detector as an indicator of when to retrain can be beneficial only if we know that the detector is not too sensitive or the detector can identify drifts. We had an example of employing a drift detector that was not able to identify any drift for the Alibaba dataset, namely the KS-ALL. In this situation, using this drift detector leads to obtaining the same ROC AUC as not retraining the model at all (Static), while slightly increasing the overall energy consumption of the model in production, since the energy consumed by the drift detector needs to be considered. We also had an example of what happens if the drift detector is too sensitive and constantly signals drifts in the experiment with the Google dataset (drift detectors KS-PCA and KS-FI). Here we observed that if a drift detector is too sensitive, the amount of energy consumed by a configuration that retrains based on drift detection becomes higher than the amount of energy consumed during periodic retraining, while sometimes (in the case of KS-FI) lowering the ROC AUC by 1\%. However, for all three datasets, there was at least one configuration involving drift detection that both reduced the consumed training energy and preserved the ROC AUC. For this reason, we argue that a drift detector can benefit both the consumed energy and the model's performance only if it is properly chosen. 



\begin{tcolorbox}[title=Finding 3, colback=lightgreen!5]
Although inference is not affected, the energy consumption of training a model is severely impacted by the retraining strategy.
\end{tcolorbox}
Throughout our experiments, we noticed that the difference in consumed energy during inference among different retraining strategies is either not significant or extremely small. This shows that inference energy consumption is in our experimental setup not influenced by the employed retraining technique. However, we solely experimented with the Random Forest classifiers since this is the state of the art of failure prediction, which is the case study of this paper. Therefore, our conclusions apply to this classification algorithm, but future work should investigate whether the same conclusions hold when employing other classification algorithms or when investigating other study cases, such as deep learning applications, which are more energy-intensive when it comes to inference.

\begin{tcolorbox}[title=Finding 4, colback=lightgreen!5]
The retraining frequency has a significant impact on the energy consumption.
\end{tcolorbox}
While answering RQ1c, we performed a one-year estimation to compare the long-term benefits of employing drift-based retraining with periodic retraining. Throughout this experiment, we also noticed that the most energy-intensive model is the failure prediction model for the Google dataset (approx. 10 megajoules consumed during one year), followed by Alibaba (approx. 1 megajoules) and Backblaze (approx. 6 kilojoules). The main difference between these models in the energy consumed over one year is the retraining frequency, since Backblaze and Google models are retrained every month and the Alibaba dataset is retrained every week. Therefore, the retraining frequency has a significant impact on the model's energy consumption. However, the retraining frequency is context dependent because it must consider the business context and how the ML application is used in real world. Hence, given the tremendous impact of the retraining frequency on the energy consumed, we recommend ML practitioners to design ML applications that require retraining less often if the business context allows.

\section{Threats to Validity}

\paragraph{External validity.} Our study focuses solely on Failure Prediction models, where concept drift is a known issue~\cite{datasplittingdecisions, towardsaconsistentinterpretation, modelmaturity}. To support replication, we provide a replication package with all the code necessary to execute experiments. All employed datasets are publicly available (Backblaze, Google, Alibaba). We use Random Forests, the state-of-the-art for this task, but results may vary with other algorithms. Due to time constraints, we limited our scope, though future work should explore other domains and models.

\paragraph{Internal validity.} Our experiments replicate state-of-the-art literature on failure prediction models~\cite{datasplittingdecisions, towardsaconsistentinterpretation, modelmaturity} and we acknowledge that different configurations can challenge the results we collect. Nevertheless, our replication package paves the work for future research to expand the scope of our subjects.

\paragraph{Construct validity} We measured energy only for training, inference, and drift detection, as other pipeline parts should not affected by retraining methods. Due to a one-week experiment budget, we ran fewer repetitions for the Google dataset, but results remain valid since 30 seeds (on Alibaba and Backblaze) showed little variation compared to 5 (used for Google).

\section{Conclusion}
In this study, we investigated the effects on training and inference energy consumption of multiple retraining techniques (sliding window vs. a full-history and periodic vs. informed). We conduct our experiments on a real-world ML application, namely failure prediction.

We provide empirical evidence that retraining a model only on the newest data (sliding window) is more sustainable than retraining a model on all available data (full-history)~\cite{mypapergreens, greentactics}. Furthermore, employing a full-history approach does not always come with a benefit in terms of the models' performance compared to sliding windows. We further showed that retraining based on unsupervised drift detectors is better than retraining periodically w.r.t. energy consumption only when the drift detection technique is not too sensitive and is capable of identifying drift. To understand the long-term benefits of employing a drift detection-based retraining technique, we showed that using an appropriate drift detector is estimated to decrease energy consumption by up to 40\% during one year compared to period retraining. Furthermore, we demonstrated that in most situations the retraining approach does not influence the energy consumed during inference.

The general conclusions derived from this study should serve as a guideline for ML practitioners to build Green ML applications. If the business context allows, practitioners should design ML applications that require less frequent retraining. For example, an ML application retrained monthly is more sustainable than retrained weekly. Practitioners should discard old data when retraining the model and should adopt a drift detection-based retraining approach to optimize energy efficiency. Unsupervised drift detectors that perform feature reduction (KS-FI and KS-PCA) are usually more sustainable than the ones that do not (KS-ALL). We recommend practitioners carefully choose the drift detector that is the most suitable for their datasets and analyze their consequences on the ML model's consumed energy and accuracy in the long term.

Based on our findings, drift detection-based retraining can solely be beneficial in terms of both energy consumption and the model's accuracy if the drift detector is properly chosen. Therefore in the future, we plan to develop a framework that enables ML practitioners to check whether a drift detector is too sensitive to data changes or incapable of detecting drifts. In this paper, we solely focused on the energy consumption of AIOps applications since these applications are widely used in industry~\cite{helenaAIOps}, they use plenty of data, and they are sensitive to concept drift~\cite{datasplittingdecisions}. We plan to expand this study to other energy-intensive applications where model update is also critical, such as large language models or other deep learning applications.


\bibliographystyle{unsrt}
\bibliography{bibliography}

\end{document}